\documentclass{article}

\usepackage{PRIMEarxiv}

\usepackage[utf8]{inputenc} 
\usepackage[T1]{fontenc}    
\usepackage{hyperref}       
\usepackage{url}            
\usepackage{booktabs}       
\usepackage{amsfonts}       
\usepackage{nicefrac}       
\usepackage{microtype}      
\usepackage{lipsum}
\usepackage{fancyhdr}
\usepackage{amsmath}
\usepackage{caption}
\usepackage{subcaption}
\usepackage{graphicx}       
\graphicspath{{media/}}     

\title{Image Generation with Multi-modal Priors using
Denoising Diffusion Probabilistic models
\thanks{\textit{Work in Progress} 
\textbf{}} 
}

\author{
  Nithin Gopalakrishnan Nair, Wele Gedara Chaminda Bandara \\
  Student Member, IEEE \\
  Johns Hopkins University \\
  Baltimore\\
  \texttt{\{ngopala2, wbandar1\}jhu.edu} \\
   \And
  Vishal M Patel \\
  Senior Member, IEEE \\
  Johns Hopkins University \\
  Baltimore\\
  \texttt{vpatel36@jhu.edu} \\
}

\begin{document}
\maketitle

\begin{abstract}
Image synthesis under multi-modal priors is a useful and challenging task that has received increasing attention in recent years. A major challenge in using generative models to accomplish this task is the lack of paired data containing all modalities (i.e. priors) and corresponding outputs. In recent work, a variational auto-encoder (VAE) model was trained in a weakly supervised manner to address this challenge. Since the generative power of VAEs is usually limited, it is difficult for this method to synthesize images belonging to complex distributions. To this end, we propose a solution based on a denoising diffusion probabilistic models to synthesise images under multi-model priors. Based on the fact that the distribution over each time step in the diffusion model is Gaussian, in this work we show that there exists a closed-form expression to the generate the image corresponds to the given modalities. The proposed solution does not require explicit retraining for all modalities and can leverage the outputs of individual modalities to generate realistic images according to different constraints. We conduct studies on two real-world datasets to demonstrate the effectiveness of our approach.
\end{abstract}

\section{Introduction}
Precise user desired images could be generated using complementary information describing the image obtained from multiple sources. Conditional image generation has received much attention in recent years for its practical utility in content creation tools\cite{huang2021multimodal}. As the number of conditions increases, the task of image synthesis becomes more constrained, yet also happens according to the user's expectations. Hence, condtional image generation there are practical utility in entertainment industry,surveillance and security, etc. Consider the problem of face synthesis, some specific conditional information, such as segmentation masks, can give a rough description of how the generated face should be, and finer details can be added by conditioning on specific text-based attributes. The task of generating images given multiple constraints is called multi-modal image synthesis. Existing methods use different generative models \cite{xia2021tedigan,huang2021multimodal,shi2019variational}(such as VAE, GAN etc.) for multimodal image synthesis. 


One major challenge for training generative models for multi-modal image synthesis is that the lack of paired data containing multiple modalities\cite{wu2018multimodal}. This is one of the primary reasons why most existing models constrain themselves to one or two modalities\cite{wu2018multimodal,shi2019variational}. Few works perform multi-modal generation with more than two modalities\cite{huang2021multimodal}. These methods are used to perform high-resolution image synthesis and needs to be trained with paired data across different domains to achieve good results. 

A weakly supervised method for multi-modal generation has been recently proposed that does not require paired data from all modalities\cite{shi2019variational}, and the model performs well when trained with sparse data. But if we need to increase the number of modalities, the model needs to be retrained, hence not being easily scalable. Scalable multi-modal generation is an area that has not been properly explored due to the difficulty in obtaining the large amounts of paired data needed to train models for the generative process. Modern multi-modal generative methods use VAEs and GANs, which lack the ability to model complex distributions.



Recently diffusion models thrived over other generative models for the task of image generation\cite{dhariwal2021diffusion}. This is due to the power of diffusion models to perform exact sampling from very complex distributions\cite{sohl2015deep}. One unique quality of diffusion models over other generative processes is that the model performs the generation through a Markovian process that happens through many timesteps. The output at each timestep is easily accessible. Hence being more flexible over other generative models and this form of generation allows manipulation of images. Multiple techniques have used this interesting property of diffusion models for low-level vision tasks like Image inpainting, Image super-resolution, and other Image restoration problems. 

In this paper, we exploit this flexible property of denoising diffusion probabilistic models and use it to design solutions to multi-modal image generation problems without explicitly retraining the network with paired data from all modalities. Specifically, we develop a derivation for computing the conditional score for the presence of multiple modes by using the conditional score for each mode. The implications of the proposed multi-modal generative method are manifold. We can easily extend multi-modal generation with complementary information across datasets using the proposed method. The labels/modalities present across can be combined for the generative problem for a particular class. Therefore, the proposed method eliminates the need for paired data for all multi-modal image generation task classes.


\section{Related Work}

\subsection{Multimodal Image generation:}
Recently multimdoal image synthesis has gained significant attention\cite{huang2021multimodal,shi2019variational,sutter2021generalized,suzuki2016joint,wu2018multimodal,xia2021tedigan,zhang2021m6}, where these method attempt to learn the posterior distribution of an image when conditioned on the prior joint distribution of all the different modalities. The approaches \cite{shi2019variational,suzuki2016joint,wu2018multimodal,zhang2021m6,sutter2021generalized} follows a variational autoencoder based solution where the all the input modalities are passed through their respective encoders to obtain the means and variances of the underlying Gaussian distributions and their means and variances in latent spaces are combined using product of experts.Finally a new sample is obtained by sampling using the new posterior means and variances. Some methods using GANs for multimodal Image synthesis have also gained recent attention. Huang et al\cite{huang2021multimodal} performs multimodal Image sysnthesis using introduces a new Local-Global Adaptive Instance normalization to combine the modalities. Huang et al \cite{huang2021multimodal} also uses Product of experts theory the combine encoded feature vectors and use a feature decoder to obtain the final output. TediGAN\cite{xia2021tedigan} uses a StyleGAN based framework where, the different visual-linguistic modalities are combined in the feature space and decoded to obtain the final output. TediGAN allows user defined Image manipulation according to the input of different modalities.

\subsection{Diffusion Probabilistic Models:}
Diffusion probabilistic models belong to a class of generative models where the a model is trained to generate images belonging to a distribution through a Markovian process. The model is trained  for reducing the lower bound of the negative log likelihood of the data distribution. Diffusion models has a forward process as well as a reverse process. In the forward process noise with very small variances are added to the image over a large number of  $T$ resulting in an isotropic Gaussian distribution. In the reverse Markovian process, a network is utilized to perform iterative denoising based starting from an isotropic Gaussian. At each step the network input is the output of the previosu step as well as a particular timestep $t$. The sampling during the forward process at a particular timestep $t$ is represented by
\begin{align}
q(z_t | z_{t-1}) &= \mathcal{N}\left(z_t ; \sqrt{1-{\beta}_{t}} z_{t-1},{\beta}_{t} \mathbf{I}\right) 
\end{align} 

Here $\beta_1,\beta_t,...\beta_t$ refers to a predefined variance schedule for the diffusion process at a timestep $t$. The corresponding reverse step is is defined by the distribution $p(z_{t-1}|z_{t}$ defined by,
\begin{align}
p(z_{t-1} | z_{t}) &= \mathcal{N}\left(\mathbf{\mu_{\theta}( z_{t},t)},\mathbf{\Sigma_{\theta}( z_{t},t)}\right) 
\end{align}

Given the samples $z_{t-1}$ and $z_t$ at timesteps $t-1$ and $t$, a neural network is used to estimate the mean of the distribution $p(z_{t-1}|z_{t})$. The solution for $\mathbf{\Sigma_{\theta}( z_{t},t)}=\sigma_t^2\mathcal{I}$ can be set to  $\mathbf{\beta_t}$ as mentioned in Ho et al\cite{ho2020denoising}.

\subsection{Conditional Denoising Diffusion Models:}
Recently Saharia et al\cite{saharia2021image} proposed a method for conditonal image generation and extended its application to image super resolution. Saharia et al\cite{saharia2021image} proposed that the conditonal generation of samples according to a particular distribution $p(z|x)$ could be performed by  adding the same condition to the denoising network with the condition $x$ at each timestep during training and inference  The conditional distribution during the reverse time process and condition with the image $x$ at each timestep. Extending this to our case, given a particular modality $x_i$ The conditonal generation process could be trained by using a newtwork with parameters $\theta$. The network $f_{\theta}(.)$ predicts the scaled version of noise $\epsilon$ in the noisy sample of the image $z_t$ . The training objective is defined as,

\begin{equation}
 L_{simple} =E_{t \sim[1, T], \epsilon \sim \mathcal{N}(0, \mathbf{I})}\left[\left\|\epsilon-f_{\theta}\left(\mathbf{z}_t,x_i, t\right)\right\|^{2}\right]\\
\end{equation}
and the sampling process takes place through the equation
\begin{equation}
    z_{t-1} \leftarrow{} \frac{1}{\sqrt{1-\beta_t}} \left( z_t - \frac{ \beta_t}{\sqrt{1-\bar{\alpha}_t}}f_{\theta}(z_t,x_i, t) \right) + \sigma_t^2 \boldsymbol{\eta}, \text{ where } \boldsymbol{\eta} = \mathcal{N}(\boldsymbol{0}, \boldsymbol{I}),
    \label{eq:cond_samp}
\end{equation}
\subsection{Langevian score based sampling from a diffusion process}

An alternate interpretation of denoising diffusion probabilistic models is denoising score based approach\cite{song2019generative}, where the sampling during inference is performed using langevian dynamics. Here a network is used to compute the score representing the gradient of likelihood of the data. The sampling during the reverse timestep is can be represented by\cite{song2019generative}
\begin{align}
    z_{t-1} \leftarrow{} \frac{1}{\sqrt{1-\beta_t}} \left( z_t -  \beta_t s_{\theta}(z_t, t) \right) + \sigma_t^2 \boldsymbol{\eta}, \text{ where } \boldsymbol{\eta} = \mathcal{N}(\boldsymbol{0}, \boldsymbol{I}),
    \label{eq:score}
\end{align}

Comparing Equations \ref{eq:cond_samp} and \ref{eq:score}, we can see that the score value 
\begin{equation}
    s_{\theta|}(z_t,t) = \frac{f_{\theta}(z_t,x_i,t)}{\sqrt{1-\bar{\alpha}_t}}
\end{equation}
\section{Proposed Method}

In normal conditional denoising diffusion models\cite{saharia2021image}, the input image is concatenated along with the sampled noise while passing through the network. When multiple modalities are present, one trivial way of finding the conditional distribution, is to concatenate all the $N$ input modalities while training the network. But In case we want to improve the functionality of the network by conditioning on new modalities, the whole model needs to be retrained by conditioning with the new $N+1$ modalities. 
 
 For a given generative task based on multimodal priors, given the $N$ modalities and models trained on them individually, the new modal for the joint generation task can be defined by 


\subsection{Multimodal conditioning using Diffusion models(MCDM)}
Let $z$ and $x_i$ denote a point in the space of the images of a particular domain and $p(z|x_i)$ denotes the distribution of the predicted image $z$ based on the modality $x_i$.Let the distribution of the image conditioned on all modalities is denoted by $P(z|x_1,x_2,...,x_N)$
and the distribution of the image conditioning on the individual modalities be $P(z|x_i)$. Assuming that all the modalities are statistically independent,

\begin{equation}
    P(z|x_1,x_2,...,x_N)=\frac{P(z)}{P(x_1,x_2,...,x_N)}\prod_{i=1}^{N}P(x_i|z) \propto \frac{\prod_{i=1}^{N}P(z|x_i)}{\prod_{i=1}^{N-1}P(z)}
    \label{eq:POE}
\end{equation}

Assuming the individual distributions $P(z|x_i)$ and $P(z)$ are Gaussian distributions, the distribution $P(z|x_1,x_2,...,x_N)$ will also be a Gaussian distribution.
Now Let's assume that $N$ diffusion models trained to generate samples the distributions $P(z|x_i)$, conditioned on each modality $x_i$ separately. 

As mentioned in section, we can use the relationship between diffusion models and score matching\cite{song2019generative} to derive the effective score function for the conditional distibution $P(z|x_1,x_2,...,x_N)$ and perform inference based on this.
\begin{equation}
     P(z_t|x_1,x_2,...,x_N)=\frac{K.\prod_{i=1}^{N}P(z|x_i)}{\prod_{i=1}^{N-1}P(z)}
     \end{equation}
     \begin{equation}
     \nabla_{z_t}log  P_{\theta}(z_t|x_1,x_2,...,x_N) =   \nabla_{z_t}log  \left(\frac{K.\prod_{i=1}^{N}P(z|x_i)}{\prod_{i=1}^{N-1}P(z)}\right)
          \end{equation}
    \begin{equation}
     = \sum_{i=1}^N \nabla_{z_t}\text{log} P(z_t|x_i)-(N-1) \nabla_{z_t}\text{log}  P(z_t)
     \end{equation}
Where K is a term independent of $z_t$. Hence the effective score when conditioned on all the modalities can be represented in terms of scores of the individual conditional distribution as well as the score of the unconditonal model. Hence the effective score $s_{c}$
\begin{align}
    s_{c} = \frac{\epsilon_c}{\sqrt{1-\bar{\alpha}_t}}
\end{align}

\begin{align}
   \epsilon_{c}= \epsilon_{\theta}(z_t,x_1,x_2,..,x_N,t) = \sum_{i=1}^N \epsilon_{\theta}(z_t,x_i,t)-(N-1)\epsilon_{\theta}(z_t,t)
\end{align}

$\epsilon_{\theta}(z_t,x_i,t)$ denotes the output prediction of the individual conditional networks and $\epsilon_{\theta}(z_t,t)$ is the prediction of the unconditional network. After computing the effective score, sampling could be performed using equation by,
\begin{equation}
    z_{t-1} \leftarrow{} \frac{1}{\sqrt{1-\beta_t}} \left( z_t - \frac{ \beta_t}{\sqrt{1-\bar{\alpha}_t}}\epsilon_c \right) + \sigma_t^2 \boldsymbol{\eta}, \text{ where } \boldsymbol{\eta} = \mathcal{N}(\boldsymbol{0}, \boldsymbol{I}),
\end{equation}




\section{Experiments}

For the multimodal generation task, we perform experiments on CelebA dataset. As the different modalities. We choose edge maps of the images as well as partial face segmentation masks from the CelebAHQ dataset. We also take the text captions for images as another modality.

\section{Results}

In this section we detail on the qualitative experiments performed. As part of the multimodal generation task, we perform three experiments combining three different modalities taking two at a time and another experiment with all three modalities.

\begin{figure}[h]
    \centering
    \begin{subfigure}[t]{0.19\linewidth}
      \captionsetup{justification=centering, labelformat=empty, font=scriptsize}
      \includegraphics[width=1\linewidth]{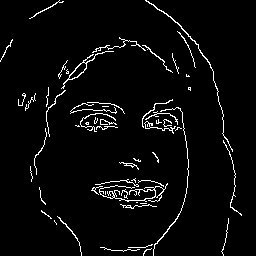}
      \includegraphics[width=1\linewidth]{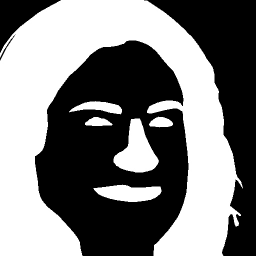}
      \includegraphics[width=1\linewidth]{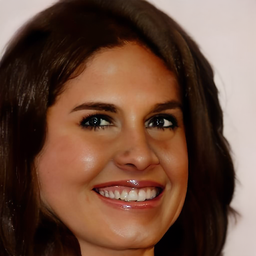}
    \end{subfigure}
    \begin{subfigure}[t]{0.19\linewidth}
      \captionsetup{justification=centering, labelformat=empty, font=scriptsize}
      \includegraphics[width=1\linewidth]{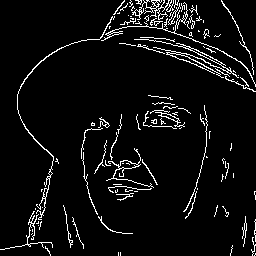}
      \includegraphics[width=1\linewidth]{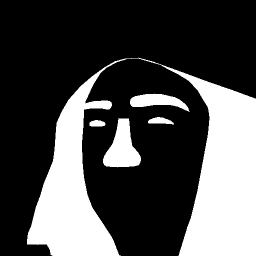}
      \includegraphics[width=1\linewidth]{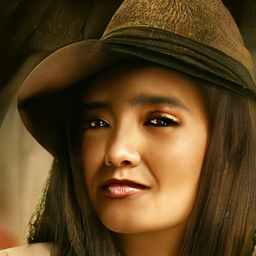}
    \end{subfigure}
    \begin{subfigure}[t]{0.19\linewidth}
      \captionsetup{justification=centering, labelformat=empty, font=scriptsize}
      \includegraphics[width=1\linewidth]{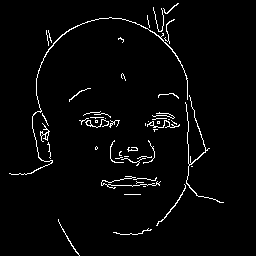}
      \includegraphics[width=1\linewidth]{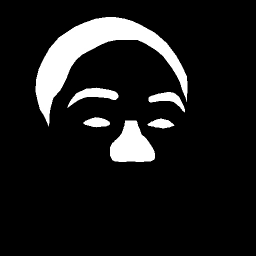}
      \includegraphics[width=1\linewidth]{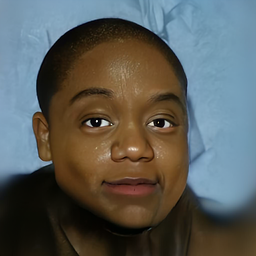}
    \end{subfigure}
    \begin{subfigure}[t]{0.19\linewidth}
      \captionsetup{justification=centering, labelformat=empty, font=scriptsize}
      \includegraphics[width=1\linewidth]{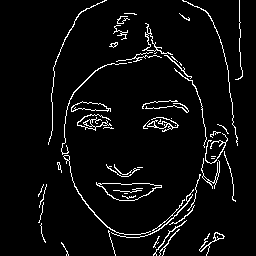}
      \includegraphics[width=1\linewidth]{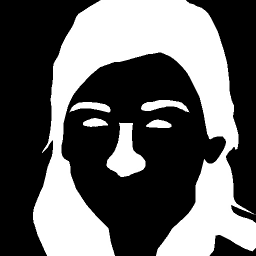}
      \includegraphics[width=1\linewidth]{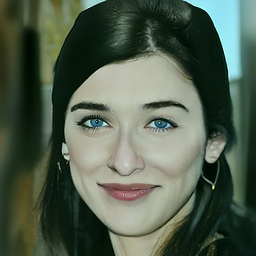}
    \end{subfigure}
    \begin{subfigure}[t]{0.19\linewidth}
      \captionsetup{justification=centering, labelformat=empty, font=scriptsize}
      \includegraphics[width=1\linewidth]{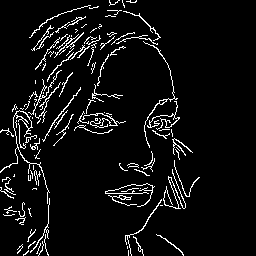}
      \includegraphics[width=1\linewidth]{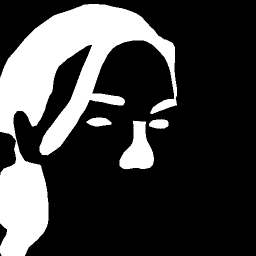}
      \includegraphics[width=1\linewidth]{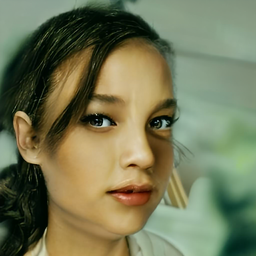}
    \end{subfigure}

    \caption{Results using the modalities sketch and partial face mask.}
    \label{fig:skseg}
  \end{figure}

\begin{figure}[h]
    \centering
    \begin{subfigure}[t]{0.19\linewidth}
      \captionsetup{justification=centering, labelformat=empty, font=scriptsize}
      \includegraphics[width=1\linewidth]{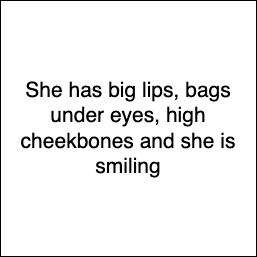}
      \includegraphics[width=1\linewidth]{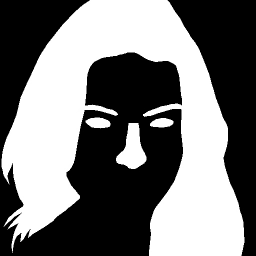}
      \includegraphics[width=1\linewidth]{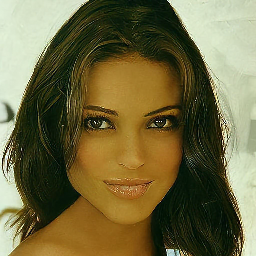}
    \end{subfigure}
    \begin{subfigure}[t]{0.19\linewidth}
      \captionsetup{justification=centering, labelformat=empty, font=scriptsize}
      \includegraphics[width=1\linewidth]{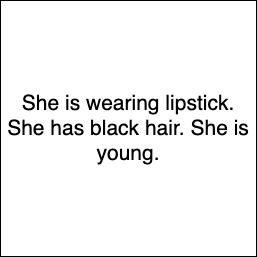}
      \includegraphics[width=1\linewidth]{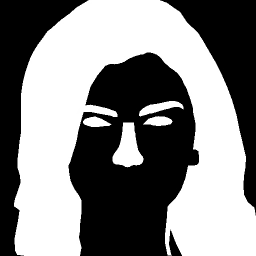}
      \includegraphics[width=1\linewidth]{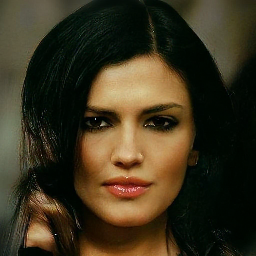}
    \end{subfigure}
    \begin{subfigure}[t]{0.19\linewidth}
      \captionsetup{justification=centering, labelformat=empty, font=scriptsize}
      \includegraphics[width=1\linewidth]{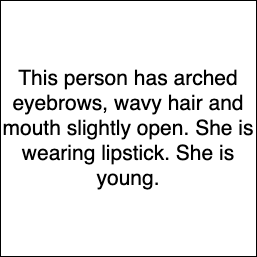}
      \includegraphics[width=1\linewidth]{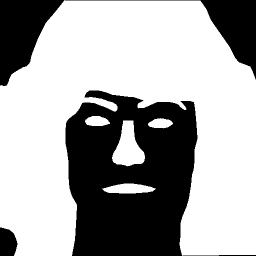}
      \includegraphics[width=1\linewidth]{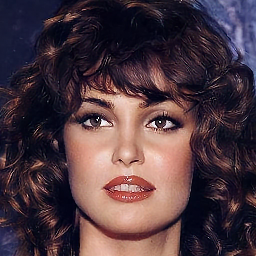}
    \end{subfigure}
    \begin{subfigure}[t]{0.19\linewidth}
      \captionsetup{justification=centering, labelformat=empty, font=scriptsize}
      \includegraphics[width=1\linewidth]{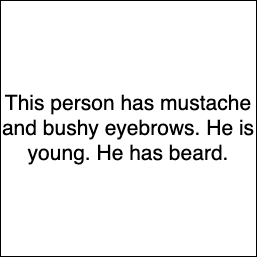}
      \includegraphics[width=1\linewidth]{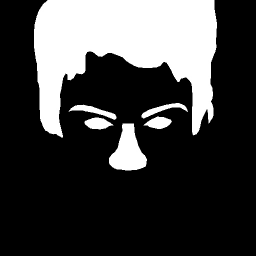}
      \includegraphics[width=1\linewidth]{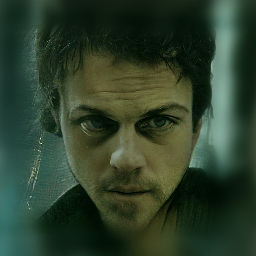}
    \end{subfigure}
    \begin{subfigure}[t]{0.19\linewidth}
      \captionsetup{justification=centering, labelformat=empty, font=scriptsize}
      \includegraphics[width=1\linewidth]{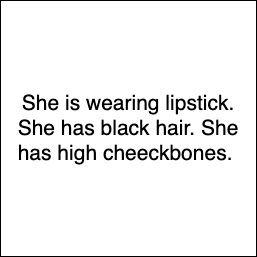}
      \includegraphics[width=1\linewidth]{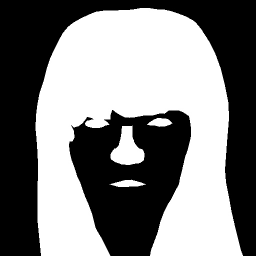}
      \includegraphics[width=1\linewidth]{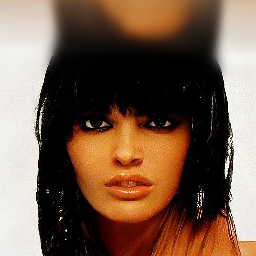}
    \end{subfigure}

   \caption{Results using the modalities text and partial face mask.}

    \label{fig:segtext}
  \end{figure}

\begin{figure}[h]
    \centering
    \begin{subfigure}[t]{0.19\linewidth}
      \captionsetup{justification=centering, labelformat=empty, font=scriptsize}
          \includegraphics[width=1\linewidth]{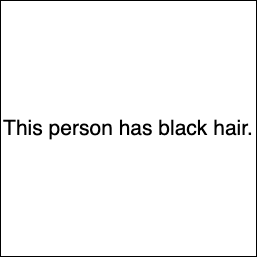}
      \includegraphics[width=1\linewidth]{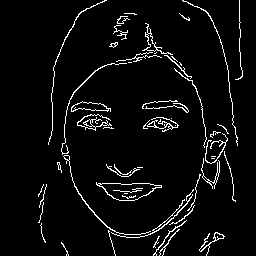}
      \includegraphics[width=1\linewidth]{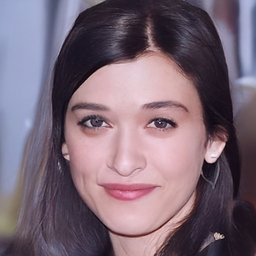}
    \end{subfigure}
    \begin{subfigure}[t]{0.19\linewidth}
      \captionsetup{justification=centering, labelformat=empty, font=scriptsize}
      \includegraphics[width=1\linewidth]{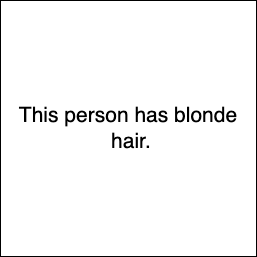}
      \includegraphics[width=1\linewidth]{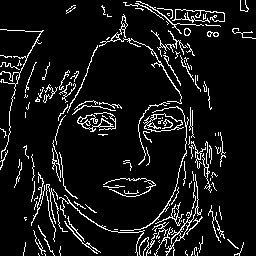}
      \includegraphics[width=1\linewidth]{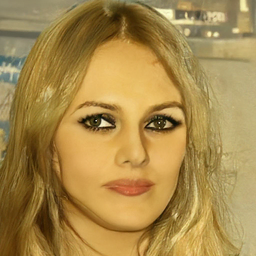}
    \end{subfigure}
    \begin{subfigure}[t]{0.19\linewidth}
      \captionsetup{justification=centering, labelformat=empty, font=scriptsize}
      \includegraphics[width=1\linewidth]{results/sktext/blonde.png}
      \includegraphics[width=1\linewidth]{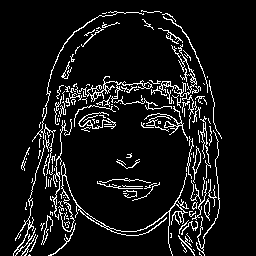}
      \includegraphics[width=1\linewidth]{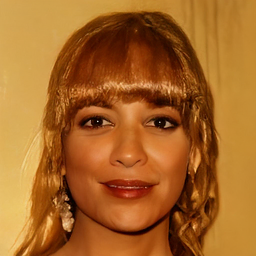}
    \end{subfigure}
    \begin{subfigure}[t]{0.19\linewidth}
      \captionsetup{justification=centering, labelformat=empty, font=scriptsize}
      \includegraphics[width=1\linewidth]{results/sktext/black.png}
      \includegraphics[width=1\linewidth]{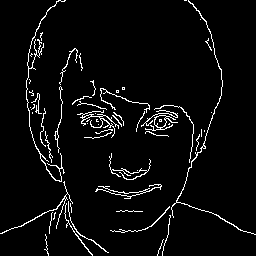}
      \includegraphics[width=1\linewidth]{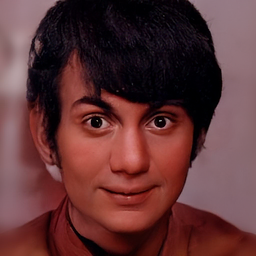}
    \end{subfigure}
    \begin{subfigure}[t]{0.19\linewidth}
      \captionsetup{justification=centering, labelformat=empty, font=scriptsize}
      \includegraphics[width=1\linewidth]{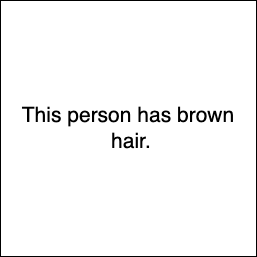}
      \includegraphics[width=1\linewidth]{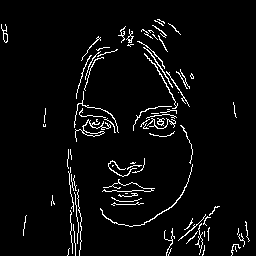}
      \includegraphics[width=1\linewidth]{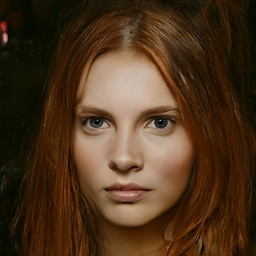}
    \end{subfigure}

    \caption{Results using the modalities sketch and text.}
.
    \label{fig:sktext}
  \end{figure}

\begin{figure}[h]
    \centering
    \begin{subfigure}[t]{0.19\linewidth}
      \captionsetup{justification=centering, labelformat=empty, font=scriptsize}
      \includegraphics[width=1\linewidth]{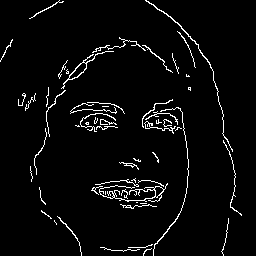}
      \includegraphics[width=1\linewidth]{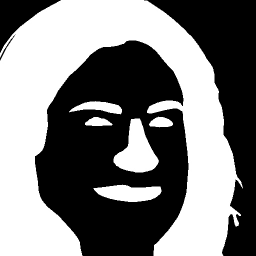}
      \includegraphics[width=1\linewidth]{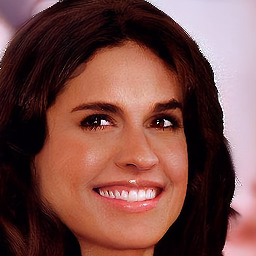}
    \end{subfigure}
    \begin{subfigure}[t]{0.19\linewidth}
      \captionsetup{justification=centering, labelformat=empty, font=scriptsize}
      \includegraphics[width=1\linewidth]{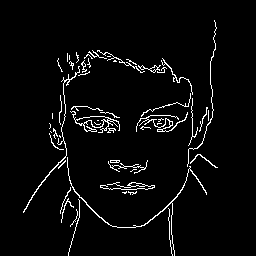}
      \includegraphics[width=1\linewidth]{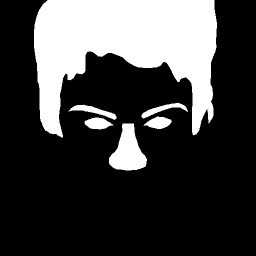}
      \includegraphics[width=1\linewidth]{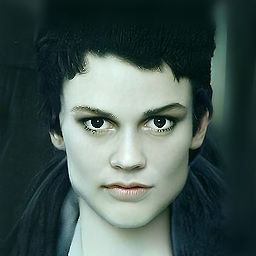}
    \end{subfigure}
    \begin{subfigure}[t]{0.19\linewidth}
      \captionsetup{justification=centering, labelformat=empty, font=scriptsize}
      \includegraphics[width=1\linewidth]{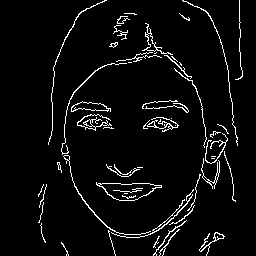}
      \includegraphics[width=1\linewidth]{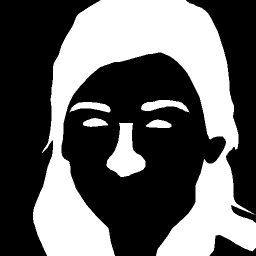}
      \includegraphics[width=1\linewidth]{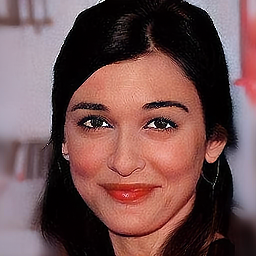}
    \end{subfigure}
    \begin{subfigure}[t]{0.19\linewidth}
      \captionsetup{justification=centering, labelformat=empty, font=scriptsize}
      \includegraphics[width=1\linewidth]{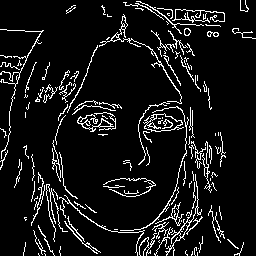}
      \includegraphics[width=1\linewidth]{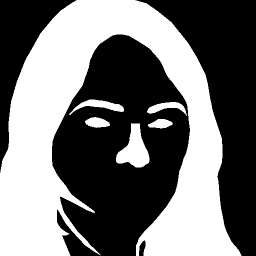}
      \includegraphics[width=1\linewidth]{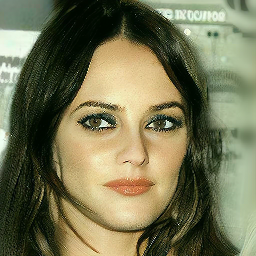}
    \end{subfigure}
    \begin{subfigure}[t]{0.19\linewidth}
      \captionsetup{justification=centering, labelformat=empty, font=scriptsize}
      \includegraphics[width=1\linewidth]{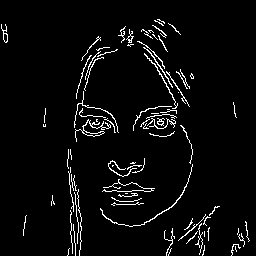}
      \includegraphics[width=1\linewidth]{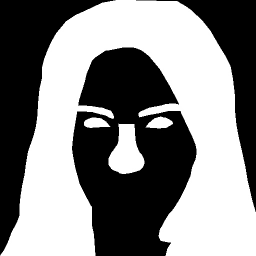}
      \includegraphics[width=1\linewidth]{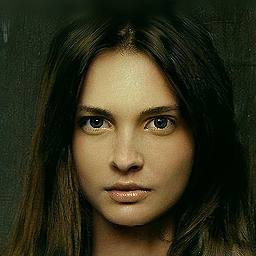}
    \end{subfigure}

    \caption{Image samples generated with three modalities sketch, segment and text. The text prompt used in all these results are \textit{"This person has black hair"}.}
    \label{fig:ALL}
  \end{figure}
  
\subsection{Modalities- sketch and segment:}

As can be seen in Figure \ref{fig:skseg} The method is able to extract complementary information across modalities and fill in the missing regions.

\subsection{Modalities- text and segment:}

As can be seen in Figure \ref{fig:segtext} Since there is a lot of missing information in the partial face masks. We are able to include key information such as hair colour, gender, mouth orientation using text prompts.

\subsection{Modalities- text and sketch:}

As can be seen in Figure \ref{fig:sktext}. The edge maps consists of a lot of information from the original image. Hence it is difficult to add much text information to this. For this reason, we choose hair colour as a property and try to change it using the corresponding text prompts.

This is a raw version of the paper. The paper will be updated with more results and experiments



\




\section{Conclusion}

In this paper we propose a method to combine multiple diffusion models during inference time to perform multimodal image synthesis. Our method does hence opens up an idea for utilizing annotated data across multiple datasets and use these together for a better constrained generation. Extensive experiments on CelebA dataset shows the working and proof of concept of our method. 
\clearpage
\bibliographystyle{unsrt}  
\bibliography{references}

\end{document}